\documentclass{article}
\usepackage{spconf,amsmath,graphicx}
\usepackage{hyperref}
\usepackage{algpseudocode}
\usepackage{url}
\usepackage[ruled,vlined]{algorithm2e}
\usepackage{tabularx}
\usepackage{multirow}
\usepackage{makecell}
\usepackage{microtype}
\usepackage{colortbl}
\usepackage[colorinlistoftodos,textwidth=20mm]{todonotes}
\usepackage{mathtools}
\usepackage{booktabs}
\usepackage{enumitem}
\usepackage{caption}
\usepackage{subcaption}
\usepackage{longtable}

\usepackage{wrapfig,lipsum,booktabs}

\newcommand{\jobid}[1]{\ignorespaces}

\def\tco{{\it train-clean-100}}
\def\tct{{\it train-clean-360}}
\def\tof{{\it train-other-500}}

\def\devclean{{\it dev-clean}}
\def\devother{{\it dev-other}}
\def\testclean{{\it test-clean}}
\def\testother{{\it test-other}}

\definecolor{grey}{cmyk}{0.1,0.1,0.1,0.1}
\definecolor{orange}{cmyk}{0.1,0.7,0.5,0.2}

\newcommand{\librispeech}{{LibriSpeech}}

\title{More speaking or more speakers?}
\name{Dan Berrebbi$^\ddagger$\sthanks{Work done during internship at Apple.}, Ronan Collobert$^\dagger$, Navdeep Jaitly$^\dagger$, Tatiana Likhomanenko$^\dagger$} 
\address{$^\ddagger$Carnegie Mellon University, $^\dagger$Apple}

\begin{document}
\maketitle
\begin{abstract}
Self-training (ST) and self-supervised learning (SSL) methods have demonstrated strong improvements in automatic speech recognition (ASR). In spite of these advances, to the best of our knowledge, there is no analysis of how the composition of the labelled and unlabelled datasets used in these methods affects the results. 
In this work we aim to analyse the effect of number of speakers in the training data on a recent SSL algorithm (wav2vec 2.0), and a recent ST algorithm (slimIPL).
We perform a systematic analysis on both labeled and unlabeled data by varying the number of speakers while keeping  the number of hours fixed and vice versa.
Our findings suggest that SSL requires a large amount of unlabeled data to produce high accuracy results, while ST requires a sufficient number of speakers in the labelled data, especially in the low-regime setting. In this manner these two approaches improve supervised learning in different regimes of data composition.

\end{abstract}
\section{Introduction and Related Works}
Self-supervised learning (SSL)~\cite{baevski2020wav2vec,hsu2021hubert,chung2021w2v,baevski2022data2vec}, self-training (ST), or pseudo-labeling (PL),~\cite{kahn2020self,xu2020iterative,likhomanenko2020slimipl,manohar2021kaizen,higuchi2021momentum,mpl2022} and their combination~\cite{xu2021self,zhang2022bigssl} have produced strong improvements in the accuracy of ASR systems when a large amount of unlabeled data is available. These works showed results on small amount of supervised data, and also studied how they scale with more unlabeled data. A recent work~\cite{riviere2021asr4real} has analysed the accuracy of open-sourced SSL and supervised models under different conditions (e.g. conversational speech) and across population subgroups (e.g. gender, accent) after training on a fixed dataset. 
In this work, we aim to understand the dependence between the number of speakers in the labeled/unlabeled training data and the accuracy of SSL/ST trained models\footnote{In this study labeled and unlabeled data come from the same domain. However unlabeled data contains not only clean speech as labeled data but also noisy speech.}. 
  
Recent work~\cite{qian2022contentvec} showed that forcing speaker-disentangled representations during SSL pre-training improves accuracy on a set of content-related downstream tasks.
Authors of~\cite{barnard2009asr} established that fewer than 50 speakers and around 10 to 20 hours of speech per language were sufficient for the development of a usable ASR system for low resource languages. 
Later wav2vec 2.0~\cite{baevski2020wav2vec} showed that ASR can be built with even 10min of supervision as long as a large unlabeled dataset is available.
Compared to~\cite{barnard2009asr} which used HMM-based models we use Connectionist Temporal Classification (CTC)~\cite{graves2006connectionist} loss with a transformer model and the latest developments in ASR training pipelines (e.g. SpecAugment~\cite{park2019specaug}). 

Earlier work~\cite{oliver2018realistic} investigated the effect of varying the amount of labeled and unlabeled data for SSL and ST in computer vision and reported that the accuracy of SSL and ST techniques tends to converge to the same value as the number of labels grows, but found surprisingly different levels of sensitivity to varying data amount across these techniques.

\begin{table}[!t]
    \caption{Proposal summary on data collection for ST and SSL. We denote $S_L$ and $S_U$ as the number of speakers and $H_L$ and $H_U$ as the total number of hours in the labeled and unlabeled data respectively. With `$*$' we mark cases where the thresholds are crucial for acceptable performance. We do not test configurations with less than 10h (100h) of labeled (unlabeled) data.} 
    \label{tab:recommendation}
    \vspace{-0.3cm}
    \begin{center}
    \resizebox{\linewidth}{!}{
    \begin{tabular}{l|rr|r}
        \toprule
        Training & \# $S_L$ & \# $S_U$ & \# $H_U$ \\
        \midrule
        wav2vec (97M) $H_L\geq10h$ & $>120$ & $>2000$  & $>500h^*$ \hspace{-2.5mm} \\
        wav2vec (317M), $H_L\geq10h$ & $>120$ & $>1000$ & $>500h^*$ \hspace{-2.5mm} \\
        \midrule
        slimIPL (255M), $H_L=10h$ & $>120^*$ \hspace{-2.5mm} & $>1000$ & $>100h$ \\
        slimIPL (255M), $H_L>10h$ & $>120$ & $>1000$ & $>100h$ \\
        \bottomrule
    \end{tabular}
    }
    \end{center}
    \vspace{-0.4cm}
\end{table}

To inform more optimal data collection, we study how SSL and ST algorithms perform as the number of speakers is varied, while keeping the number of hours fixed and vice versa for both the labeled and the unlabeled datasets. 
A prior work~\cite{sanabria2022measuring} provides insights for dependence on speaker diversity in SSL trained only with synthesized data and in limited configurations.
To the best of our fknowledge, we are the first to perform a systematic survey focused on the role of the number of speakers for SSL and ST methods.
Our findings encourage re-thinking of the data collection process for ASR, especially in a low-resource supervision and lead us to propose an appropriate composition which is summarized in Table~\ref{tab:recommendation}.
We highlight the main results from~Table~\ref{tab:recommendation} as: i) for low-resource settings, it is critical to have a sufficient number of speakers in labeled data; ii) the improvement from increasing the number of speakers in both labeled and unlabeled data plateaus after a certain threshold; iii) it is critical for SSL to have enough unlabeled data, while for ST it is critical to have enough speakers in the labeled data.

\section{Experimental Setup}
\subsection{Generation of Data Subsets} All our experiments were performed using \librispeech{}~\cite{panayotov2015librispeech}. 
{\bf Labeled sets:} We took \tco{} with 100h and 251 speakers and sampled different number of speakers and hours. {\bf Unlabeled sets:} We took \tct{} and \tof{} with 860h and 2087 speakers in total and sample different number of speakers and hours. Note, there is no overlap in speakers between \librispeech{} provided sets (train, dev or test). 

\begin{table}[!ht]
    \caption{Statistics on speakers extracted from \tco{} (100h), \tct{} (360h) and \tof{} (500h).} 
    \label{tab:speakers}
    \vspace{-0.3cm}
    \begin{center}
    \resizebox{\linewidth}{!}{
    \begin{tabular}{rc|rc|rc}
        \toprule
        \multicolumn{2}{c}{100h} & \multicolumn{2}{c}{360h} & \multicolumn{2}{c}{500h} \\ 
        \midrule
         \#speakers & Female \% & \#speakers & Female \% & \#speakers & Female \%\\
        \midrule
        24 & 54.1 & 115 & 48.7 & 145 & 48.3 \\
         60 & 53.3 & 230 & 46.5 & 290 & 50.7 \\
         120 & 51.6 & 460 & 47.2 & 580 & 49.3 \\
          185 & 52.4 & 921 & 47.7 & 1166 & 48.4\\
         251 & 49.8 & - & - & - & - \\
        \bottomrule
    \end{tabular}
    }
    \end{center}
\end{table}

\begin{table}[!t]
    \caption{Statistics of labeled (randomly extracted from \tco{}) and unlabeled (randomly extracted from \tct{} and \tof{}) subsets of \librispeech{} with different amount of speakers and hours.} 
    \vspace{-0.3cm}
    \label{tab:speakers-sets}
    \setlength\tabcolsep{5pt} 
    \begin{center}
    \resizebox{\linewidth}{!}{
    \begin{tabular}{lrrrrrr}
        \toprule
        \multirow{2}{*}{Name} & \multirow{2}{*}{\#hours} & \multirow{2}{*}{\#speakers} & \#min per & \multirow{2}{*}{\#utterances} & \multicolumn{2}{c}{\#words} \\ \cmidrule{6-7}
         &  &  & speaker & & Total & Unique \\
        \midrule
        10h-24 & 10.1 & 24 & 25.0 & 2846  & 97355 & 12108\\
        10h-60 & 9.9 & 60 &  10.0 & 2836  & 95772 & 12860\\
        10h-120 & 9.8 & 120 & 5.0 & 2793  & 96015 & 12969\\
        10h-185 & 9.5 & 185 & 3.2 & 2709  & 93466 & 12897\\
        10h-251 & 9.6 & 251 & 2.4 & 2728  & 94469 & 13098\\
        \midrule
        25h-60 & 25.1 & 60 & 25.0 & 7172  & 243251 & 20989\\
        25h-120 & 24.8 & 120 & 12.5 & 7050  & 243367 & 21114\\
        25h-185 & 24.7 & 185 & 8.1 & 6986  & 241711 & 21444\\
        25h-251 & 24.6 & 251 & 6.0 & 7017  & 243180 & 21705\\
        \midrule
        50h-120 & 50.3 & 120 & 25.0 & 14282  & 492905 & 29546\\
        50h-185 & 49.7 & 185 & 16.2 & 14046  & 486793 & 30008\\
        50h-251 & 49.6 & 251 & 12.0  & 14109  & 488881 & 30418\\
        \midrule 
        100h-251 & 100.6 & 251 & 24.0 & 28539  & 990101 & 41572\\
        \midrule
        \midrule
        108h-260 & 110.6 & 260 & 25.0 & 32410  & 1078816 & 43286\\
        108h-520 & 107.4 & 520 & 12.5 & 31438  & 1051319 & 44639\\
        108h-1040 & 105.5 & 1040 & 6.2 & 31007  & 1026921 & 45863\\
        108h-2087 & 103.9 & 2087 & 3.1 & 30951  & 1017510 & 46809\\
        \midrule
        216h-520 & 221.3 & 520 & 25.0 & 64633  & 2165301 & 60360\\
        216h-1040 & 214.8 & 1040 & 12.5  & 62895  & 2091514 & 62186\\
        216h-2087 & 211.5 & 2087 & 6.2 & 62803  & 2071413 & 63737\\
        \midrule
        430h-1040 & 442.5 & 1040 & 25.0 & 129419  & 4307931 & 82906\\
        430h-2087 & 423.4 & 2087 & 12.4  & 125090  & 4147820 & 84694\\
        \midrule
        860h-2087 & 860.5 & 2087 & 24.7 & 252702 & 8413454 & 162561\\
        \bottomrule
    \end{tabular}
    }
    \end{center}
\end{table}

We describe below a sampling strategy for labeled subsets; the same sampling strategy is applied to sample from unlabeled subsets.
To achieve fair and consistent comparisons, we implemented a sampling process satisfying the following conditions.
First, the speakers subsets are randomly sampled from \tco{}: smaller subsets of speakers are included in any larger subset of speakers.
In Table~\ref{tab:speakers} we show statistics on the sampled speakers gender: percentage of female and male speakers is well balanced across our sampled subsets. Second, for every speaker we sample different amount of minutes preserving the inclusion of subsets per speaker: e.g. if we have two subsets containing 2min and 10min for the same speaker, then the 2min audio samples are included in the 10min ones. 
In \librispeech{} every speaker has at most 25min of audio, thus some combinations cannot be generated, e.g. 25h with 24 speakers. The summary of all subsets is detailed in Table~\ref{tab:speakers-sets}.

\subsection{Models and Algorithms}
For this study we run various models with three different training algorithms -- supervised training, continuous pseudo-labeling with slimIPL~\cite{likhomanenko2020slimipl} and SSL with wav2vec 2.0~\cite{baevski2020wav2vec}. 

\subsubsection{Supervised Training and Self-Training}
For both supervised training and ST with slimIPL we closely follow slimIPL~\cite{likhomanenko2020slimipl} and use the same architecture and training pipeline. 
Models are trained with English letters token set\footnote{26 letters augmented with the apostrophe and a word boundary token.}, CTC loss, identical SpecAugment~\cite{park2019specaug} parameters, and Adagrad optimizer~\cite{duchi2011adaptive}.
The acoustic model is the same transformer model introduced in slimIPL, except that we encode positions with a recently proposed CAPE~\cite{likhomanenko2021cape} instead of relative positional embedding~\cite{shaw2018self} to speed up training by 2-3x and decrease the memory footprint significantly. All models (255M parameters) are trained on 8 GPUs of A100-40GB for a maximum of 500-600k updates. We use dynamic batches that pack $\sim 290$s of audio per GPU. For supervised training we use dropout and layer drop~\cite{fan2019reducing} of 0.3 for 50h, 0.4 for 25h and 0.5 for 10h, while for slimIPL we use 0.3 for 108h, 0.25 for 216h, 0.2 for 430h, and 0.1 for 860h of unlabeled data. The slimIPL cache size is set to 1000 (which was the optimal cache size for 10h and was a robust setting for 100h of supervision) with cache probability of 0.1. We additionally search over the proportion of labeled/unlabeled data $\lambda$ varying it in $\{1, 2, 5, 10\}$ and the burn in period before pseudo-labels are used (after 20k, 30k, 40k steps) for best results.

\subsubsection{Self-Supervised Training}
For wav2vec~2.0~\cite{baevski2020wav2vec} we use open-sourced recipes from \texttt{fairseq}\footnote{\tiny \url{https://github.com/facebookresearch/fairseq/blob/main/examples/wav2vec/}} for both Base (95M) and Large (317M) models. 
First, we pre-train on our unlabeled subsets with different number of hours and speakers: we use the \librispeech{} pre-training recipe and only change the encoder dropout for different amount of pre-training hours: 0.05 (0.2) for 860h, 0.1 (0.3) for 430h, 0.15 (0.4) for 216h, and 0.2 (0.5) for 108h for a Base (Large) model. The remaining parameters are unchanged. Pre-trained models are fine-tuned on different labeled subsets using the 10h recipe for all 10h/25h subsets and the 100h recipe for 50h/100h subsets using the same parameters as in the open-sourced recipe.

\subsubsection{Evaluation}
We evaluate models on standard \librispeech{} sets, \devclean{}, \devother{}, \testclean{}, \testother{}. For the sake of space we report only dev sets in all Figures. For supervised training we run experiments with three different seeds and report mean and standard deviation of the word error rate (WER). For ST and SSL experiments we run only one seed and report its WER on \devother{} set.

\section{Results}

\subsection{Impact of Dataset Composition on Supervised Training}
\begin{figure}[ht!]
    \centering
    \includegraphics[width=0.5\textwidth]{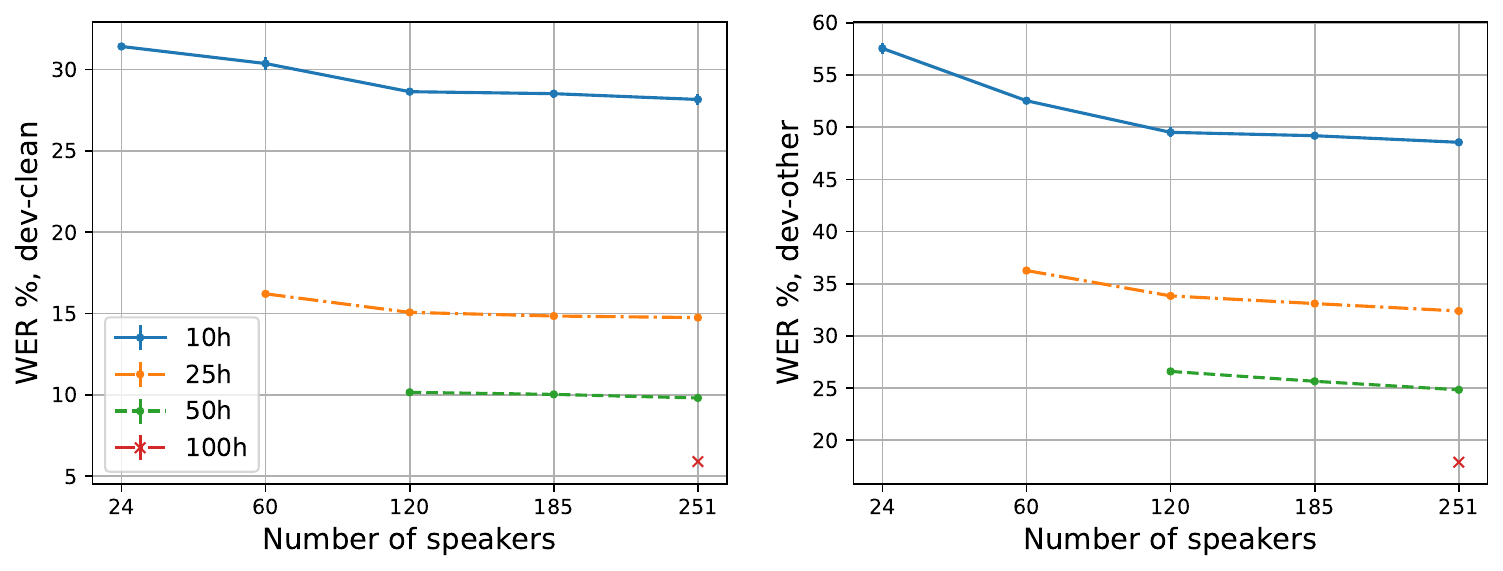}
    \caption{Dependence of supervised training on the amount of data and speakers.}
    \label{fig:sup-trend}
\end{figure}

Figure~\ref{fig:sup-trend} shows dependence between speakers and number of hours for supervised training. Clearly, more speakers is helpful for the training but at some point the gain diminishes (185 speakers).

\subsection{Impact of Dataset Composition on ST}

Figure~\ref{fig:trend} (row 1, column 3) shows the heatmap of WER for different configurations of number of speakers in labeled and unlabeled data for slimIPL. 
Increasing the number of speakers is beneficial until certain point in both labeled and unlabeled data. 
Row 4, column 3 shows that in the low-supervision setting having a large number of speakers in labeled data is critical. For example, with only 10 hours of labeled data, having 24 speakers results in a WER of 28\% (with 100h unlabeled data) and 19\% (with 860h unlabeled data) but having 120 speakers reduces the WER to 23\% and 15.5\% respectively. Row 5, column 3, on the other hand shows that the number of speakers in the unlabeled set does not make as much of a difference (it is possible that this becomes more important when the number of speakers is much smaller than the 260 we started with). 
Row 6, column 3 shows the expected trend that increasing the number of labeled/unlabeled hours produces gains in accuracy\footnote{Note that each point on the grid of the heatmap reflects averages over multiple runs with different, but fixed amount of training data. However, manually inspecting subsets confirmed that the trends look similar for settings with fixed, individual number of training hours.}.

\subsection{Impact of Dataset Composition on SSL}
\begin{figure*}[ht!]
    \centering
    \includegraphics[scale=0.35]{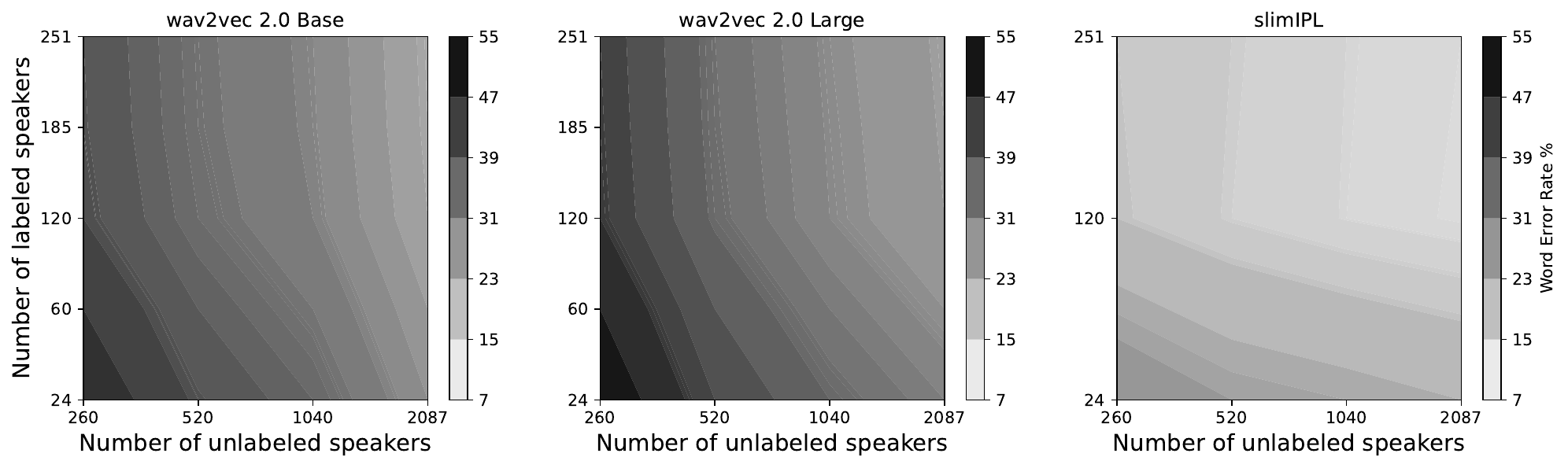}
    \includegraphics[scale=0.35]{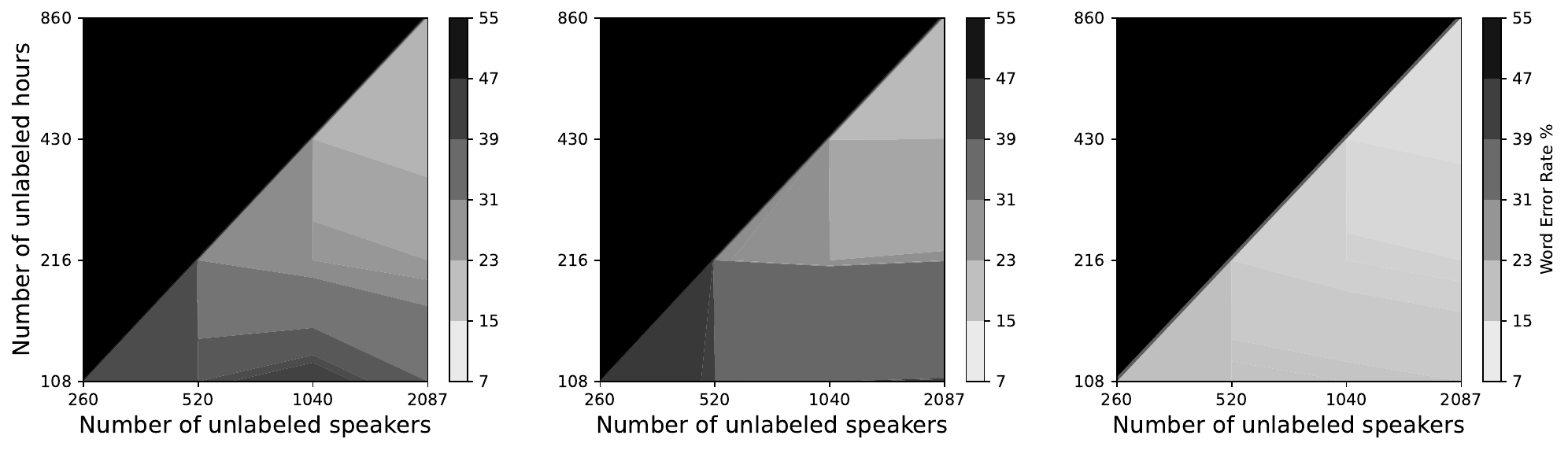}
    \includegraphics[scale=0.35]{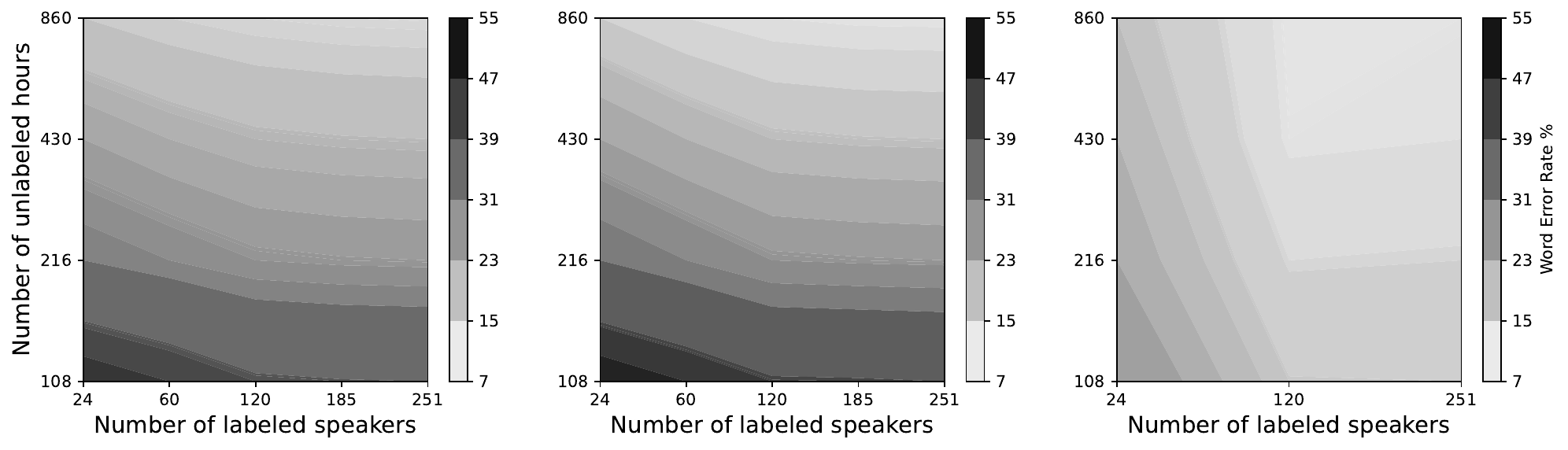}
    \includegraphics[scale=0.35]{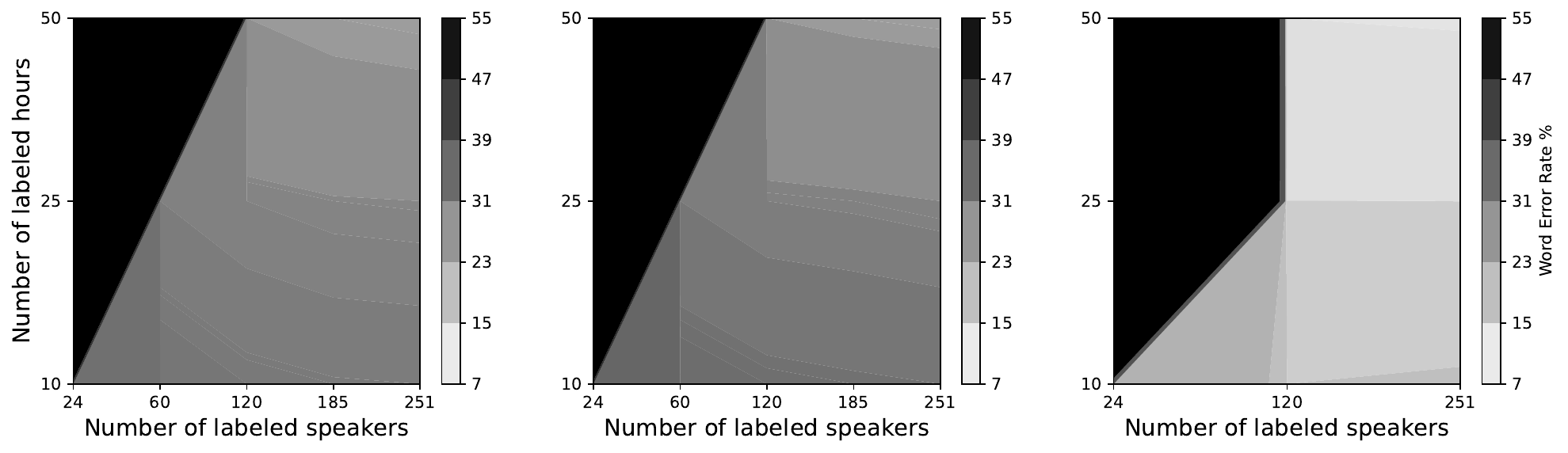}
    \includegraphics[scale=0.35]{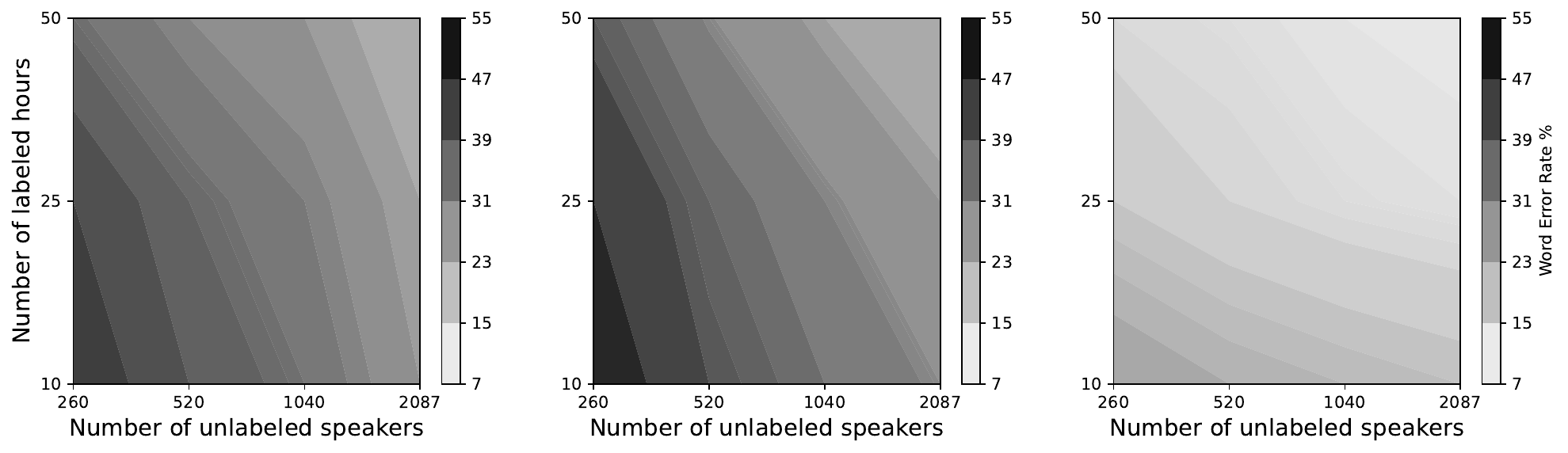}
    \includegraphics[scale=0.35]{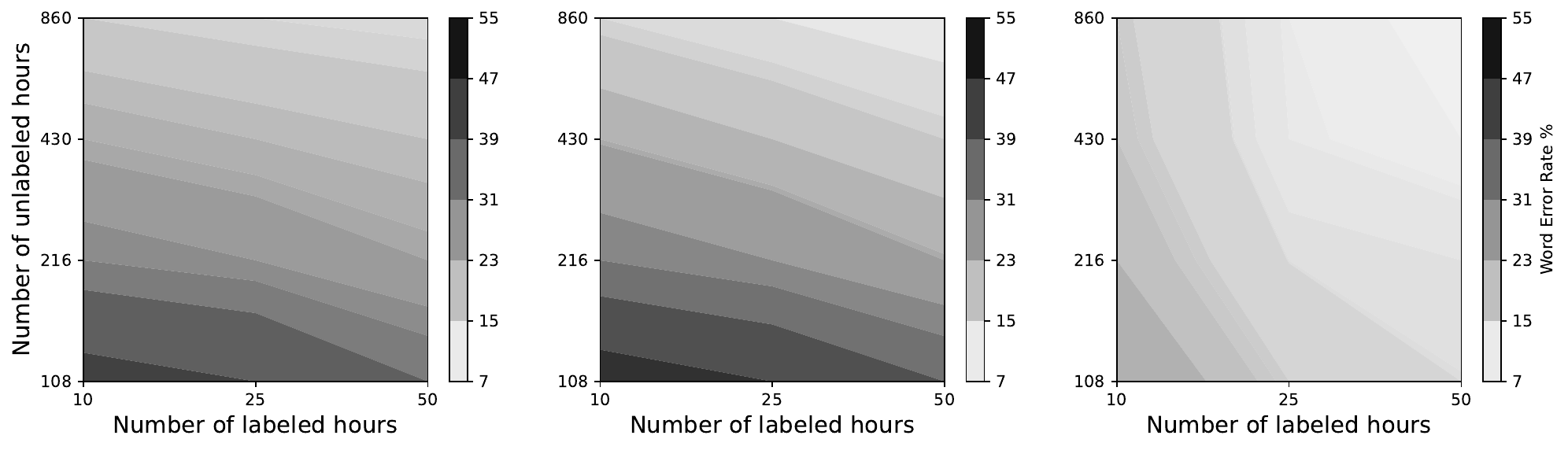}
    \caption{Word error rate (WER) heatmaps on \librispeech{} \devother{} for SSL with wav2vec 2.0 Base (column 1) and Large (column 2), and ST with slimIPL (column 3). We consider WER as a function of 4 variables: unlabeled hours, unlabeled speakers, labeled hours and labeled speakers, and plot heatmap for every pair of variables while we average over the other two variables. The black part corresponds to missing configurations for which we do not have enough data in \librispeech{}. As seen in Table~\ref{tab:recommendation} we observe that i) for low-resource settings, it is critical to have a sufficient number of speakers in labeled data; ii) the improvement from increasing the number of speakers in both labeled and unlabeled data plateaus after a certain threshold; iii) it is critical for SSL to have enough unlabeled data, while for ST it is critical to have enough speakers in labeled data.}
    \label{fig:trend}
\end{figure*}

From Figure~\ref{fig:trend}, columns 1, 2, rows 3 and 4 we see that both Base and Large wav2vec 2.0 models show small improvements with more speakers in the labeled data. 
Columns 1, 2, rows 1 and 2, likewise show small improvements from more speakers in unlabeled data.  In contrast, from columns 1, 2, row 6, we see a drastic gain in accuracy from increasing number of unlabeled hours (e.g., the large model achieves a WER of 49\% with 10h of labeled data and 100h of unlabeled data, averaged over runs with different number of speakers; this number reduces to 23\% with 400h of unlabeled data averaged over runs with different number of speakers). 

\subsection{Comparing ST and SSL}
For SSL, the number of speakers in labeled data is not critical, whereas it is essential for ST, especially in low data regimes. In contrast, the amount of unlabeled data is critical for SSL models and not as critical for ST models.
For both types of algorithms, the low-supervision regime is different from high-supervision regime and the influence of number of speakers quickly diminishes in both labeled and unlabeled data once a sufficient number of speakers is assembled.

\section{Discussion \& Conclusion}
Our observations in self-training and self-supervised learning in ASR suggest that data collection may be revisited in low supervision to improve results in ASR. 
For self-training it is better to have more speakers in labeled data, when the labeled data size is fixed. It is also beneficial to have around 1K speakers in the unlabeled data.  SSL on the other hand is affected more by the number of hours of unlabeled data than it is by the number of speakers in the unlabeled data.
In this manner these two
approaches improve supervised learning in different regimes
of dataset composition, which can explain the reason of their combination success~\cite{xu2021self,zhang2022bigssl}.
Our paper does not address the question of how to select from accents, lexical content and noise conditions for improving the performance of ST and SSL and we leave this for future research.

\clearpage

\clearpage

\begin{figure*}[h!]
    \centering
    \includegraphics[scale=0.35]{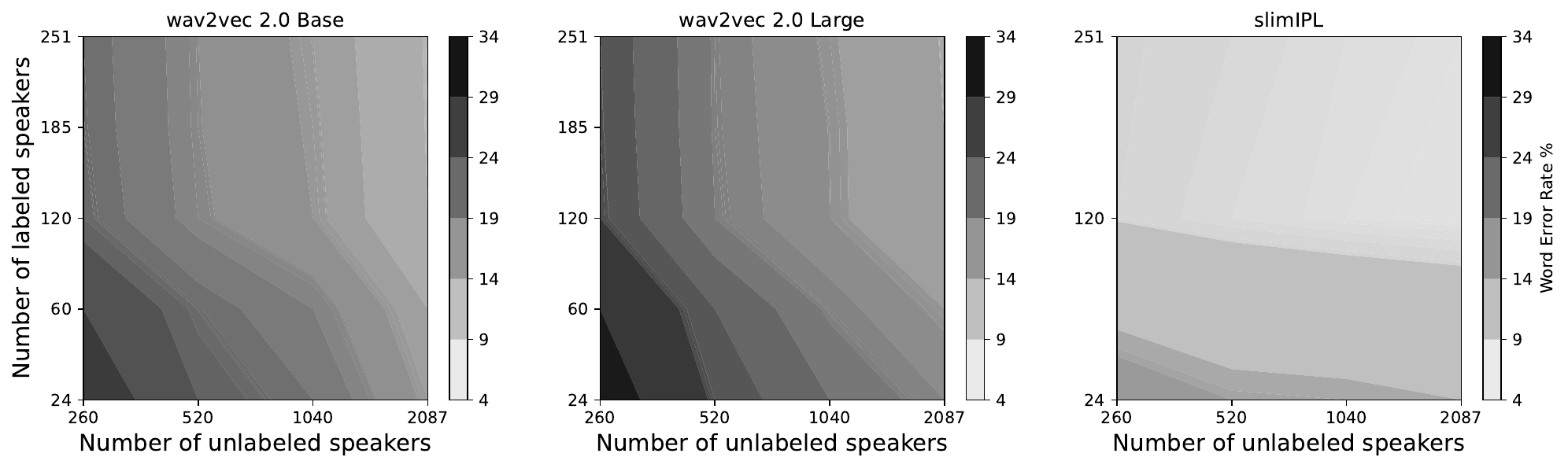}
    \includegraphics[scale=0.35]{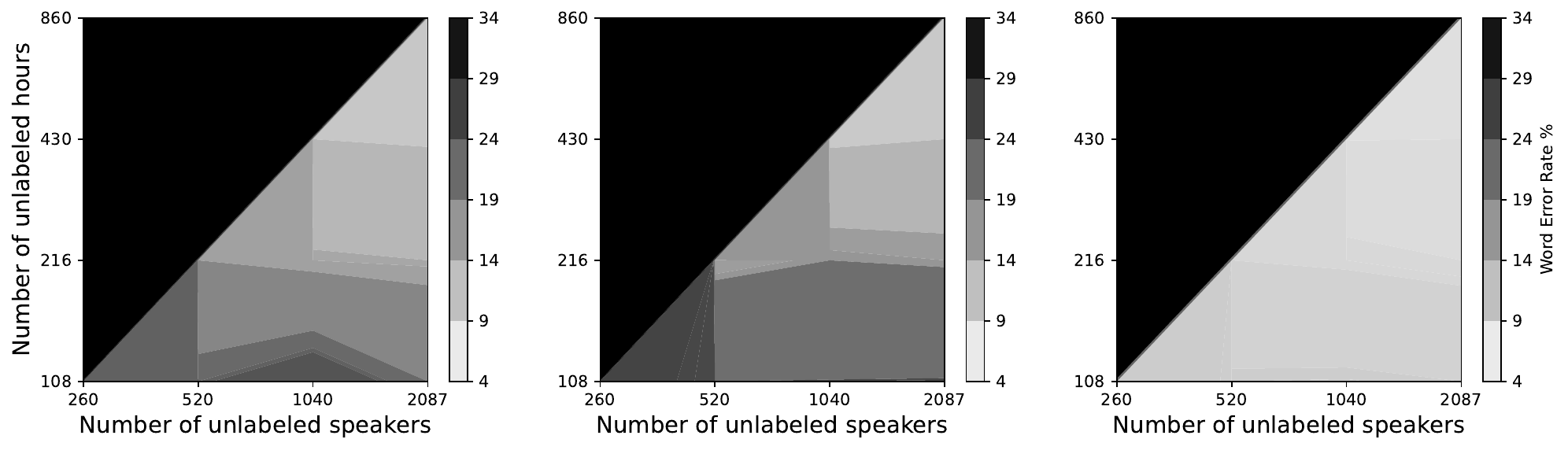}
    \includegraphics[scale=0.35]{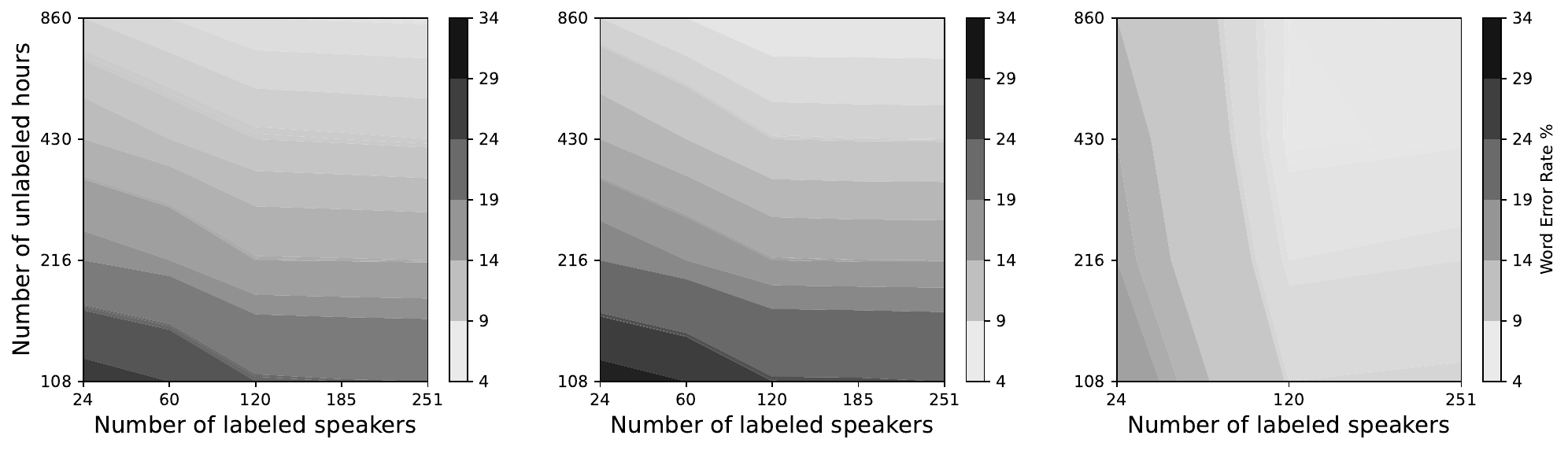}
    \includegraphics[scale=0.35]{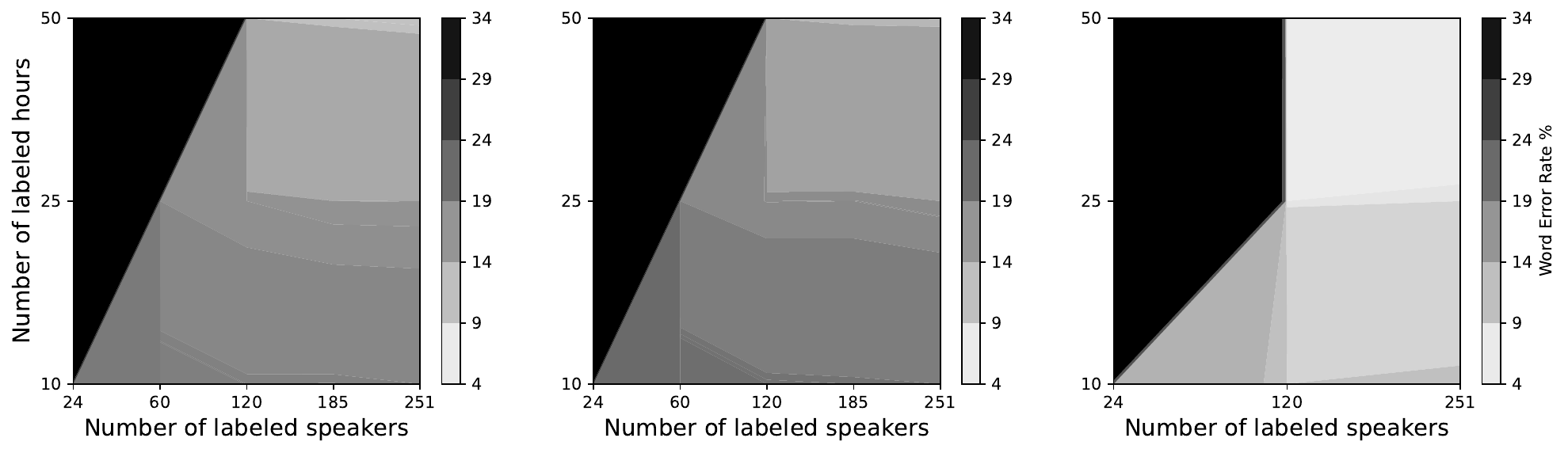}
    \includegraphics[scale=0.35]{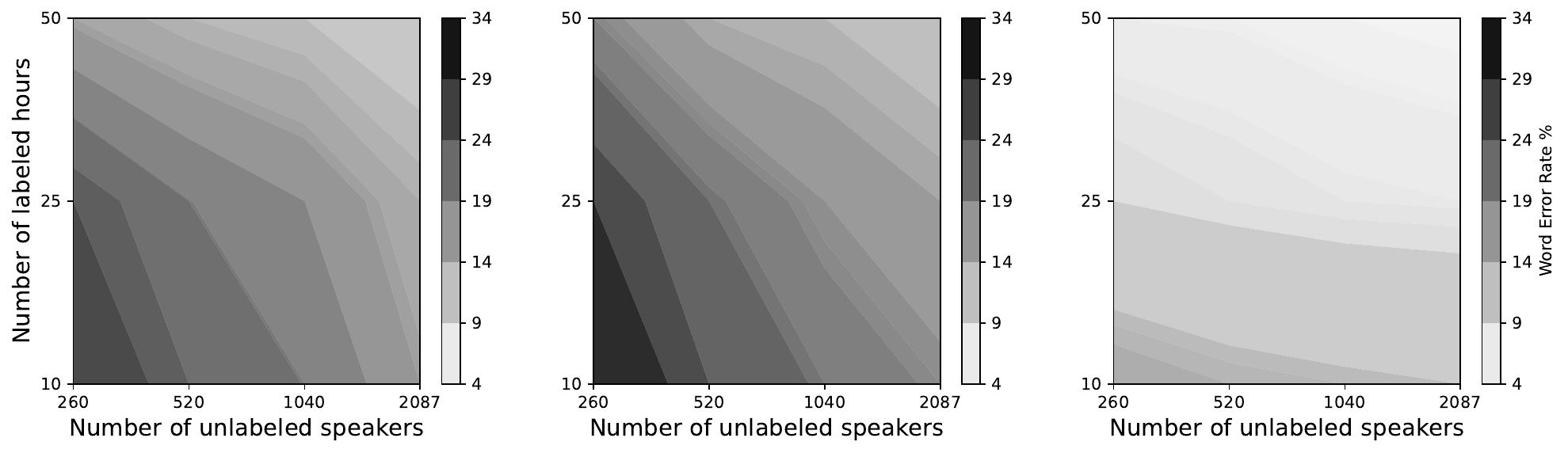}
    \includegraphics[scale=0.35]{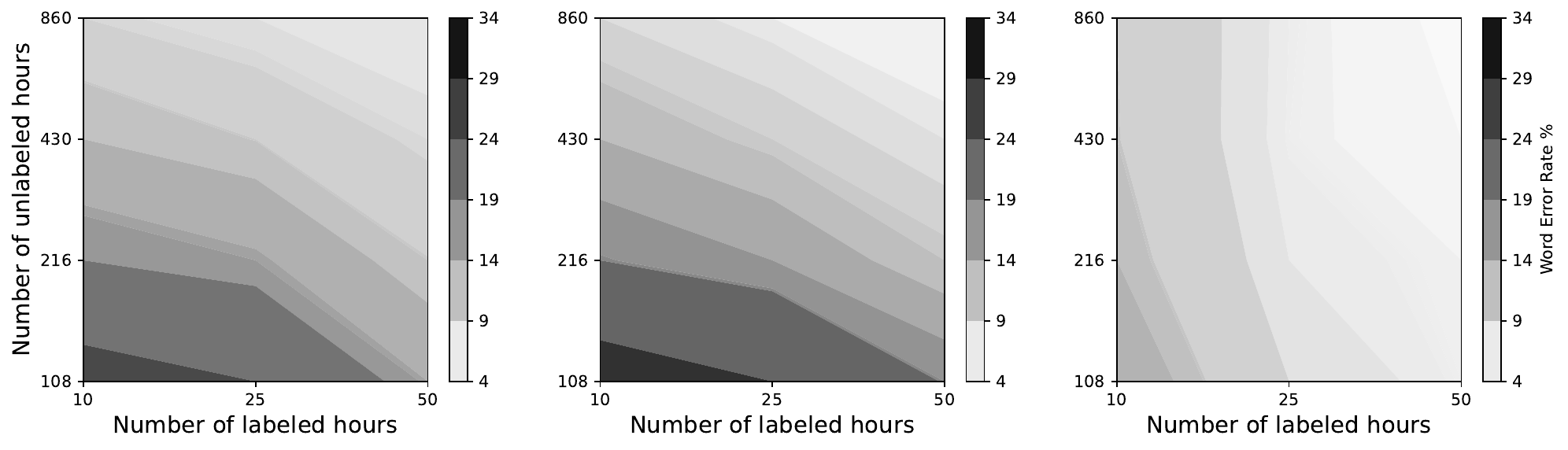}
    \caption{Word error rate (WER) heatmaps on \librispeech{} \devclean{} for SSL with wav2vec 2.0 Base (column 1) and Large (column 2), and ST with slimIPL (column 3). We consider WER as a function of 4 variables: unlabeled hours, unlabeled speakers, labeled hours and labeled speakers, and plot heatmap for every pair of variables while we average over the other two variables. The black part corresponds to missing configurations for which we do not have enough data in \librispeech{}. As seen in Table~\ref{tab:recommendation} we observe that i) for low-resource settings, it is critical to have a sufficient number of speakers in labeled data; ii) the improvement from increasing the number of speakers in both labeled and unlabeled data plateaus after a certain threshold; iii) it is critical for SSL to have enough unlabeled data, while for ST it is critical to have enough speakers in labeled data.}
    \label{fig:trend-clean}
\end{figure*}

\clearpage

\begin{table*}[!t]
    \caption{Detailed Word Error Rate (WER) for all models trained on different subsets of labeled and unlabeled data from Table~\ref{tab:speakers-sets}. Figures~\ref{fig:trend} and~\ref{fig:trend-clean} are shown based on these results.} 
    \vspace{-0.3cm}
    \label{tab:final-wer}
    \setlength\tabcolsep{7pt} 
    \begin{center}
    \resizebox{0.8\linewidth}{!}{
    \begin{tabular}{llrr|rr|rr}
        \toprule
        \multirow{2}{*}{Unlabeled data} & \multirow{2}{*}{Labeled data} & \multicolumn{2}{c}{wav2vec 2.0 Base WER \%} & \multicolumn{2}{c}{wav2vec 2.0 Large WER \%} & \multicolumn{2}{c}{slimIPL WER \%}\\ 
        \cmidrule{3-4} \cmidrule{5-6} \cmidrule{7-8}
         &  &  dev-clean & dev-other & dev-clean & dev-other & dev-clean & dev-other\\
        \midrule
\multirow{12}{*}{108h-260} & 10h-24 & 28.2 & 47.4 & 32.3 & 52.6 & 16.8 & 28.3 \\
 & 10h-60 & 27.1 & 45.4 & 31.8 & 50.6 & - & - \\
 & 10h-120 & 26.4 & 44.3 & 30.8 & 49.7 & 12.9 & 23.2 \\
 & 10h-185 & 26.2 & 44.1 & 30.4 & 49.2 & - & - \\
 & 10h-251 & 25.9 & 43.9 & 30.2 & 48.9 & 13.2 & 23.0 \\
 & 25h-60 & 24.4 & 42.7 & 27.4 & 46.5 & - & - \\
 & 25h-120 & 23.8 & 41.7 & 26.3 & 45.0 & 8.0 & 17.7 \\
 & 25h-185 & 23.3 & 41.2 & 2.06 & 44.7 & - & - \\
 & 25h-251 & 23.4 & 41.1 & 25.8 & 44.4 & 8.3 & 17.4 \\
 & 50h-120 & 15.2 & 35.4 & 18.2 & 38.1 & 6.2 & 14.4 \\
 & 50h-185 & 14.9 & 34.1 & 18.1 & 37.6 & - & - \\
 & 50h-251 & 14.7 & 33.7 & 18.0 & 37.2 & 6.0 & 14.0 \\
 \midrule
\multirow{12}{*}{108h-520} & 10h-24 & 27.2 & 46.0 & 30.7 & 49.1 & 15.5 & 26.4 \\
 & 10h-60 & 26.7 & 44.3 & 32.2 & 47.5 & - & - \\
 & 10h-120 & 25.9 & 43.4 & 29.2 & 46.6 & 12.6 & 21.9 \\
 & 10h-185 & 25.9 & 43.0 & 29.0 & 46.2 & - & - \\
 & 10h-251 & 25.8 & 43.1 & 28.8 & 46.0 & 13.3 & 22.4 \\
 & 25h-60 & 24.2 & 41.6 & 26.9 & 44.5 & - & - \\
 & 25h-120 & 23.5 & 40.6 & 26.1 & 43.4 & 8.0 & 16.4 \\
 & 25h-185 & 22.9 & 40.2 & 25.9 & 42.9 & - & - \\
 & 25h-251 & 22.9 & 40.2 & 25.8 & 42.6 & 8.2 & 16.3 \\
 & 50h-120 & 15.1 & 34.5 & 17.3 & 36.0 & 6.0 & 13.7 \\
 & 50h-185 & 14.5 & 33.4 & 16.9 & 35.3 & - & - \\
 & 50h-251 & 14.3 & 32.6 & 16.9 & 34.7 & 6.0 & 13.5 \\
 \midrule
\multirow{12}{*}{108h-1040} & 10h-24 & 31.2 & 50.3 & 30.7 & 49.3 & 16.1 & 26.7 \\
 & 10h-60 & 30.6 & 48.6 & 30.3 & 47.5 & - & - \\
 & 10h-120 & 29.8 & 47.7 & 29.6 & 46.4 & 12.8 & 22.2 \\
 & 10h-185 & 29.7 & 47.3 & 29.6 & 46.1 & - & - \\
 & 10h-251 & 29.3 & 47.2 & 29.3 & 45.9 & 13.0 & 21.5 \\
 & 25h-60 & 27.4 & 45.2 & 26.4 & 43.8 & - & - \\
 & 25h-120 & 26.8 & 44.5 & 25.9 & 42.5 & 7.8 & 15.7 \\
 & 25h-185 & 26.3 & 44.0 & 26.3 & 42.9 & - & - \\
 & 25h-251 & 26.1 & 43.8 & 25.2 & 42.1 & 8.1 & 15.9 \\
 & 50h-120 & 16.4 & 36.5 & 18.3 & 36.0 & 6.0 & 13.2 \\
 & 50h-185 & 16.0 & 35.5 & 17.9 & 35.7 & - & - \\
 & 50h-251 & 15.6 & 35.0 & 17.9 & 35.3 & 6.0 & 13.0 \\
 \midrule
\multirow{12}{*}{108h-2087} & 10h-24 & 25.2 & 41.8 & 30.9 & 49.6 & 15.4 & 25.6 \\
 & 10h-60 & 24.8 & 40.3 & 30.4 & 47.8 & - & - \\
 & 10h-120 & 24.2 & 39.8 & 29.5 & 46.8 & 12.8 & 21.1 \\
 & 10h-185 & 24.3 & 39.6 & 29.6 & 46.9 & - & - \\
 & 10h-251 & 24.3 & 39.5 & 29.4 & 46.5 & 12.9 & 21.5 \\
 & 25h-60 & 22.6 & 37.8 & 26.5 & 43.9 & - & - \\
 & 25h-120 & 21.8 & 37.0 & 25.5 & 42.7 & 7.7 & 15.6 \\
 & 25h-185 & 21.6 & 36.5 & 26.5 & 43.3 & - & - \\
 & 25h-251 & 21.7 & 36.5 & 25.3 & 42.2 & 8.0 & 15.6 \\
 & 50h-120 & 13.6 & 30.8 & 18.7 & 36.6 & 6.0 & 13.1 \\
 & 50h-185 & 13.2 & 29.9 & 18.3 & 36.2 & - & - \\
 & 50h-251 & 13.0 & 29.4 & 18.3 & 35.9 & 6.1 & 12.8 \\
 \midrule
 \multirow{12}{*}{216h-520} & 10h-24 & 18.4 & 33.7 & 18.6 & 33.6 & 14.3 & 23.5 \\
 & 10h-60 & 18.2 & 32.5 & 17.7 & 31.8 & - & - \\
 & 10h-120 & 17.7 & 31.8 & 17.2 & 30.9 & 10.7 & 18.4 \\
 & 10h-185 & 17.7 & 31.4 & 17.2 & 30.5 & - & - \\
 & 10h-251 & 17.6 & 31.4 & 17.2 & 30.4 & 11.5 & 19.7 \\
 & 25h-60 & 16.6 & 30.7 & 15.4 & 29.2 & - & - \\
 & 25h-120 & 16.0 & 30.1 & 14.8 & 28.1 & 6.7 & 14.0 \\
 & 25h-185 & 15.9 & 29.5 & 14.8 & 27.8 & - & - \\
 & 25h-251 & 15.9 & 29.4 & 14.8 & 27.7 & 6.8 & 13.8 \\
 & 50h-120 & 10.9 & 26.6 & 10.5 & 24.3 & 5.8 & 13.0 \\
 & 50h-185 & 10.9 & 25.5 & 10.3 & 23.4 & - & - \\
 & 50h-251 & 10.9 & 25.1 & 10.3 & 23.1 & 6.0 & 13.1 \\
        \bottomrule
    \end{tabular}
    }
    \end{center}
\end{table*}

\begin{table*}[!t]
    \caption{(Continue) Detailed Word Error Rate (WER) for all models trained on different subsets of labeled and unlabeled data from Table~\ref{tab:speakers-sets}. Figures~\ref{fig:trend} and~\ref{fig:trend-clean} are shown based on these results.} 
    \vspace{-0.3cm}
    \label{tab:final-wer}
    \setlength\tabcolsep{7pt} 
    \begin{center}
    \resizebox{0.8\linewidth}{!}{
    \begin{tabular}{llrr|rr|rr}
        \toprule
        \multirow{2}{*}{Unlabeled data} & \multirow{2}{*}{Labeled data} & \multicolumn{2}{c}{wav2vec 2.0 Base WER \%} & \multicolumn{2}{c}{wav2vec 2.0 Large WER \%} & \multicolumn{2}{c}{slimIPL WER \%}\\ 
        \cmidrule{3-4} \cmidrule{5-6} \cmidrule{7-8}
         &  &  dev-clean & dev-other & dev-clean & dev-other & dev-clean & dev-other\\
        \midrule
\multirow{12}{*}{216h-1040} & 10h-24 & 17.6 & 31.2 & 18.6 & 32.5 & 14.4 & 23.1 \\
 & 10h-60 & 16.8 & 29.6 & 18.0 & 31.2 & - & - \\
 & 10h-120 & 16.5 & 29.2 & 17.5 & 30.3 & 10.6 & 18.0 \\
 & 10h-185 & 16.6 & 29.1 & 17.5 & 30.2 & - & - \\
 & 10h-251 & 16.5 & 28.9 & 17.4 & 29.8 & 11.2 & 18.5 \\
 & 25h-60 & 15.6 & 28.1 & 15.7 & 28.5 & - & - \\
 & 25h-120 & 14.9 & 27.4 & 15.1 & 27.5 & 6.7 & 13.3 \\
 & 25h-185 & 14.7 & 26.9 & 15.0 & 27.1 & - & - \\
 & 25h-251 & 14.6 & 26.8 & 15.0 & 27.0 & 6.7 & 13.3 \\
 & 50h-120 & 10.5 & 24.8 & 10.6 & 23.1 & 5.8 & 12.5 \\
 & 50h-185 & 10.2 & 23.8 & 10.6 & 22.5 & - & - \\
 & 50h-251 & 9.9 & 23.7 & 10.3 & 22.1 & 5.8 & 12.4 \\
 \midrule
\multirow{12}{*}{216h-2087} & 10h-24 & 16.8 & 29.0 & 19.7 & 33.5 & 13.9 & 22.7 \\
 & 10h-60 & 16.3 & 28.3 & 19.3 & 31.8 & - & - \\
 & 10h-120 & 16.0 & 27.4 & 19.0 & 31.2 & 10.4 & 16.8 \\
 & 10h-185 & 16.4 & 27.3 & 18.7 & 31.0 & - & - \\
 & 10h-251 & 16.1 & 27.3 & 18.7 & 30.7 & 10.8 & 17.8 \\
 & 25h-60 & 14.6 & 26.3 & 16.6 & 29.0 & - & - \\
 & 25h-120 & 14.4 & 25.6 & 16.1 & 28.0 & 6.6 & 13.2 \\
 & 25h-185 & 14.3 & 25.5 & 15.9 & 27.7 & - & - \\
 & 25h-251 & 14.3 & 25.0 & 15.9 & 27.5 & 6.7 & 12.9 \\
 & 50h-120 & 10.5 & 22.8 & 11.2 & 23.3 & 5.7 & 12.2 \\
 & 50h-185 & 10.2 & 22.2 & 11.1 & 23.0 & - & - \\
 & 50h-251 & 10.4 & 21.8 & 11.0 & 22.8 & 5.7 & 12.3 \\
 \midrule
\multirow{12}{*}{430h-1040} & 10h-24 & 12.7 & 25.6 & 13.2 & 24.4 & 13.2 & 20.9 \\
 & 10h-60 & 12.0 & 24.0 & 12.7 & 22.9 & - & - \\
 & 10h-120 & 11.9 & 23.3 & 12.1 & 22.1 & 9.7 & 16.0 \\
 & 10h-185 & 11.7 & 23.0 & 12.1 & 21.9 & - & - \\
 & 10h-251 & 11.9 & 22.8 & 12.1 & 21.7 & 9.9 & 16.8 \\
 & 25h-60 & 10.8 & 22.5 & 10.2 & 20.3 & - & - \\
 & 25h-120 & 107 & 21.6 & 9.7 & 19.5 & 6.0 & 11.4 \\
 & 25h-185 & 10.3 & 21.3 & 9.6 & 19.1 & - & - \\
 & 25h-251 & 10.2 & 21.3 & 9.5 & 18.9 & 6.2 & 12.0 \\
 & 50h-120 & 8.8 & 20.4 & 7.2 & 16.7 & 5.0 & 10.4 \\
 & 50h-185 & 8.8 & 19.8 & 7.1 & 16.2 & - & - \\
 & 50h-251 & 8.2 & 19.3 & 6.9 & 15.9 & 4.9 & 10.1 \\
 \midrule
\multirow{12}{*}{430h-2087} & 10h-24 & 12.2 & 23.5 & 13.4 & 24.6 & 13.7 & 21.0 \\
 & 10h-60 & 11.9 & 22.4 & 13.1 & 23.0 & - & - \\
 & 10h-120 & 11.8 & 21.7 & 12.8 & 22.5 & 9.3 & 15.2 \\
 & 10h-185 & 11.5 & 21.4 & 12.8 & 22.1 & - & - \\
 & 10h-251 & 11.6 & 21.3 & 12.8 & 21.9 & 9.9 & 16.0 \\
 & 25h-60 & 10.6 & 20.8 & 10.6 & 20.3 & - & - \\
 & 25h-120 & 10.3 & 20.0 & 10.3 & 19.6 & 6.1 & 11.2 \\
 & 25h-185 & 10.2 & 19.6 & 10.1 & 19.1 & - & - \\
 & 25h-251 & 10.3 & 19.6 & 10.1 & 19.1 & 6.1 & 11.2 \\
 & 50h-120 & 8.5 & 18.8 & 7.4 & 16.4 & 5.0 & 10.1 \\
 & 50h-185 & 8.3 & 18.1 & 7.3 & 15.8 & - & - \\
 & 50h-251 & 8.0 & 17.8 & 7.4 & 15.7 & 4.9 & 10.2 \\
 \midrule
\multirow{12}{*}{860h-2087} & 10h-24 & 9.2 & 17.3 & 8.9 & 16.3 & 12.1 & 18.8 \\
 & 10h-60 & 8.8 & 16.4 & 8.6 & 15.1 & - & - \\
 & 10h-120 & 8.7 & 15.9 & 8.4 & 14.9 & 10.0 & 15.6 \\
 & 10h-185 & 8.6 & 15.5 & 8.5 & 14.5 & - & - \\
 & 10h-251 & 8.7 & 15.4 & 8.4 & 14.5 & 10.4 & 16.3 \\
 & 25h-60 & 7.9 & 15.3 & 6.7 & 13.0 & - & - \\
 & 25h-120 & 7.7 & 14.7 & 6.5 & 12.7 & 6.2 & 11.1 \\
 & 25h-185 & 7.6 & 14.5 & 6.4 & 12.2 & - & - \\
 & 25h-251 & 7.6 & 14.3 & 6.4 & 12.1 & 6.2 & 11.1 \\
 & 50h-120 & 6.5 & 14.3 & 4.9 & 10.7 & 4.5 & 9.5 \\
 & 50h-185 & 6.5 & 13.6 & 4.9 & 10.3 & - & - \\
 & 50h-251 & 6.2 & 13.4 & 4.9 & 10.2 & 4.6 & 9.5 \\
        \bottomrule
    \end{tabular}
    }
    \end{center}
\end{table*}

\clearpage

\bibliographystyle{IEEEbib}

\end{document}